\documentclass[11pt,a4paper]{article}

\usepackage{algorithmic}
\usepackage{graphicx}
\usepackage{textcomp}
\usepackage{xcolor}
\usepackage{hyperref}
\usepackage{booktabs}
\usepackage{multirow}
\usepackage{geometry}
\usepackage{amsmath}
\usepackage{amssymb}
\usepackage{algorithm}
\title{Loop-Extrusion Linkage: Spectral Ordering and Interval-Based Structure Discovery for Continuous Optimization}

\author{
  Eren Unlu \\
  \textit{Globeholder} \\
  Paris, France \\
  \texttt{\href{https://orcid.org/0000-0001-5380-6305}{ORCID: 0000-0001-5380-6305}}
}

\begin{document}
\maketitle

\begin{abstract}
The rapid growth of nature-inspired metaheuristics has exposed a persistent gap between metaphorical novelty and genuine algorithmic advancement. Motivated by the biophysics of chromatin loop extrusion---a well-characterized genome-folding process driven by SMC motor complexes and conditional barriers---we introduce the Loop-Extrusion Linkage (LEL) operator, a structure-learning wrapper that combines online variable-interaction estimation, spectral seriation via the Fiedler vector, and adaptive interval-based subspace search. LEL constructs a sparse interaction graph from successful optimization steps, derives a heuristic one-dimensional variable ordering, and generates overlapping evaluation subsets through stochastic interval growth modulated by learned boundary-crossing probabilities. We evaluate LEL on six synthetic diagnostic functions at $d{=}96$ designed to probe specific structural hypotheses---contiguous blocks, permuted blocks, overlapping windows, banded chains, separable controls, and dense rotated couplings---across $10^4$ and $5 \cdot 10^4$ evaluation budgets with 15 independent seeds. Results are assessed via the Wilcoxon signed-rank test with Holm--Bonferroni correction and Vargha--Delaney $\hat{A}_{12}$ effect sizes. At $10^4$ evaluations, Full LEL achieves the best median log-gap on 3 of 6 functions (contiguous and permuted block Rosenbrock, dense ellipsoid), significantly outperforming all ablations and jSO on the structured tasks. At $5 \cdot 10^4$ evaluations, simpler ablations and baselines often surpass the full method, indicating that the adaptive barrier mechanism may over-constrain late-stage search on uniformly partitioned landscapes. The strongest supported finding is that learned spectral ordering consistently and substantially improves over graph-only grouping and random variable ordering, suggesting that interaction-graph seriation is the most valuable component of the proposed framework. These results position LEL as a promising early-budget structural operator whose barrier and queueing components require further refinement.
\end{abstract}

\section{Introduction}
\label{sec:intro}

The design of continuous black-box optimization algorithms frequently requires balancing exploitation of learned problem structure against general algorithmic robustness. For non-separable or partially separable large-scale problems, discovering and preserving variable linkage---the subset of variables that interact strongly---within a restricted evaluation budget can yield meaningful efficiency gains. However, the recent expansion of nature-inspired metaheuristics has been dominated by metaphorical rebranding of generic operators (e.g., differential mutation, swarm velocity updates, or generalized random walks) rather than the introduction of structurally novel mechanisms for variable grouping or subset learning \cite{CamachoVillalon2026, CamachoVillalon2023, Sorensen2015, Latorre2021, Molina2025, Hu2024}. Extensive critical analyses have shown that many ostensibly novel algorithms remain largely isomorphic to classical Differential Evolution (DE) or Particle Swarm Optimization (PSO) once the metaphorical veneer is removed \cite{Wang2025, Ma2023, Molina2020}. To advance structured continuous optimization beyond this plateau, new methods should introduce algorithmically distinct mechanisms for discovering and exploiting variable interactions, rather than relabelling existing search dynamics \cite{Rani2024}.

In this work, we introduce the Loop-Extrusion Linkage (LEL) operator, which takes structured inspiration from the biophysics of genome folding. In eukaryotic cells, chromatin is organized through \emph{loop extrusion} driven by SMC macromolecular complexes (cohesin and condensin), which bind to the one-dimensional DNA chain and processively pull it into growing loops until arrested by probabilistic structural barriers such as CTCF sites or MCM complexes \cite{Davidson2021, Mirny2021, Dequeker2022}. This process partitions a massive linear sequence into functional domains using only localized, context-dependent boundary signals. We abstract one specific idea from this biophysics: an ordered substrate combined with partially permeable boundaries may support localized, overlapping domain discovery \cite{Banigan2020, Corsi2023}. It is important to note that, while the biological process motivates the design, the resulting algorithmic choices are engineering decisions justified by their empirical performance rather than by direct biological correspondence.

The central challenge that motivates this abstraction is computational. Traditional model-based continuous paradigms, notably the Covariance Matrix Adaptation Evolution Strategy (CMA-ES) \cite{Hansen2016}, capture full variable linkage globally through a covariance matrix, incurring $O(d^2)$ storage and $O(d^3)$ update costs that rapidly exhaust evaluation budgets in high-dimensional problems. Separable variants such as sep-CMA-ES reduce these costs to $O(d)$ by assuming diagonal covariance, but sacrifice the ability to model cross-variable dependencies \cite{Omidvar2022a}. Decomposition-based approaches such as Differential Grouping (DG2) \cite{Omidvar2017} partition variables into non-overlapping groups using finite-difference probes, then optimize each group independently via cooperative coevolution. These methods handle separable and near-separable structure well, but their grouping step is typically performed once as a preprocessing phase and does not adapt online. LEL pursues a different trade-off: rather than modeling global linkage or assuming separability, it learns a \emph{one-dimensional variable ordering} from a sparse interaction graph, then applies localized interval-based subspace search along that ordering. This approach scales efficiently because the interaction graph is maintained online, the ordering step uses a single eigenvector computation, and the variation operators act on small local blocks.

LEL performs four operations, each loosely motivated by the biology but justified on its own algorithmic merits: (1)~it estimates a sparse variable-interaction graph from successful historical search steps using exponential-moving-average weighted correlations; (2)~it \emph{seriates} this graph into a heuristic one-dimensional variable ordering using the Fiedler vector of the graph Laplacian \cite{Atkins1998}, which places strongly interacting variables in adjacent positions; (3)~across this ordering, it maintains probabilistic boundary-crossing weights that modulate where intervals can expand, adapting the boundary strengths from evidence of cross-boundary versus within-boundary improvements; and (4)~it \emph{extrudes} overlapping evaluation subsets by stochastic bidirectional interval growth, with utility-based conflict resolution when intervals compete for shared variables. The underlying variation operator is a conventional combination of elite attraction, differential mutation, and Gaussian noise (DE/rand/1/bin), deliberately chosen to be standard so that any performance differences can be attributed to the structural discovery mechanism rather than to the mutation rule.

To evaluate whether these mechanisms contribute genuine value, we employ a targeted synthetic diagnostic design at $d{=}96$ with six functions engineered to probe specific structural hypotheses: contiguous blocks (S2), permuted blocks (S3), overlapping windows (S4), banded chains (S5), and two negative controls---separable sphere (S1) and dense rotated ellipsoid (S6). We systematically ablate LEL's components---removing graph seriation, removing barriers, removing queueing---and compare against jSO \cite{Brest2017} and DG2+CC \cite{Omidvar2017}. Results are evaluated at two budgets ($10^4$ and $5 \cdot 10^4$) using the Wilcoxon signed-rank test with Holm--Bonferroni correction and Vargha--Delaney $\hat{A}_{12}$ effect sizes. Our findings indicate that learned spectral ordering is the most strongly supported component, consistently improving over graph-only and random-order baselines on structured problems. However, the full barrier and queueing mechanisms show mixed evidence: at $10^4$ evaluations Full LEL wins 3 of 6 functions, but at $5 \cdot 10^4$ simpler ablations and established baselines typically surpass it. We report these results as a mechanism study that identifies which components contribute and where further work is needed.

\section{Related Work}
\label{sec:related}

\subsection{Metaheuristic Design and Evaluation Standards}

The evolutionary and swarm intelligence literature has seen rapid growth in bio-inspired method proposals, with recent surveys cataloguing more than 500 named algorithms claiming inspiration from diverse natural phenomena \cite{Ma2023, Rani2024}. However, rigorous methodological critiques have established that this expansion is predominantly driven by metaphorical rebranding rather than fundamentally novel algorithmic operators \cite{Molina2020, Sorensen2015, CamachoVillalon2023, CamachoVillalon2026}. Detailed empirical taxonomy reviews confirm that many named algorithms---such as those inspired by wolf packs, moth navigation, or bat echolocation---share identical core mutation, crossover, and selection dynamics with established classical methods such as DE, PSO, or Evolution Strategies once the metaphorical naming is stripped away \cite{Hu2024, Wang2025}. This observation has motivated a broad push toward evaluating search algorithms on the basis of their \emph{algorithmic mechanisms} rather than their rhetorical packaging, with standards emphasizing strict hyperparameter protocols, avoidance of benchmark suites that fail to penalize structurally blind operators, and performance profiling across multiple evaluation budgets \cite{Latorre2021, Vermetten2024, Piotrowski2025, Hansen2021, More2009, Piotrowski2023}. LEL is designed to contribute a \emph{specific structural mechanism}---interaction-graph seriation followed by interval-based subspace search---rather than a complete metaphor-driven optimizer. Its evaluation targets mechanism isolation through ablation rather than broad competitive benchmarking \cite{Yang2020, Molina2025}.

\subsection{Linkage Learning and Variable Grouping}

LEL's structural objective intersects with the lineage of empirical linkage learning algorithms that seek to discover and exploit variable dependencies during optimization. The Linkage Tree Genetic Algorithm (LTGA) \cite{Thierens2010} builds dependency hierarchies through mutual-information-based clustering to regulate block-level recombination in combinatorial domains. Direct and conditional empirical linkage discovery techniques (DLED, cDLED) \cite{Przewozniczek2021, Przewozniczek2025} detect pairwise and higher-order variable dependencies from observed fitness changes, extending linkage learning to multi-structured and continuous problems. In large-scale continuous optimization, Cooperative Coevolution with Differential Grouping (DG2) \cite{Omidvar2014, Omidvar2017} identifies variable groups through finite-difference interaction probes and optimizes each group independently \cite{Omidvar2022a, Omidvar2022b}. DG2 is the most widely used decomposition method in the large-scale continuous benchmarking community and serves as one of our primary baselines. More recent approaches, including variable-interaction-graph methods for PSO \cite{Czworkowski2025} and gray-box optimization techniques that exploit known or partially known problem structure \cite{Whitley2016, Bouter2020}, infer interaction structure from historical fitness trajectories or from analytical problem representations.

LEL differs from these methods in one specific way: rather than forming static non-overlapping groups from the learned interaction graph, it first \emph{seriates} the graph into a one-dimensional linear ordering and then generates potentially overlapping intervals along that ordering as optimization subproblems. This distinction is the central hypothesis we evaluate experimentally---that linearizing the interaction graph and searching along the resulting ordering may provide a useful representation for problems whose variable dependencies have approximately serializable structure.

\subsection{Spectral Seriation}

The conversion of a pairwise similarity or interaction matrix into a one-dimensional ordering is a classical problem known as seriation. Spectral approaches based on the Fiedler vector---the eigenvector corresponding to the second-smallest eigenvalue of the graph Laplacian---are well-studied heuristics for this problem \cite{Atkins1998}. Under specific structural assumptions (e.g., the interaction matrix has Robinson or pre-R structure after permutation), spectral seriation provides exact recovery guarantees for the underlying linear order. Outside those restricted regimes, the Fiedler ordering remains a practical heuristic that tends to place strongly connected nodes near each other, without formal optimality guarantees for arbitrary graphs. LEL uses spectral seriation as a practical order-recovery tool, not as a provably optimal procedure. The quality of the recovered ordering depends on the extent to which the true variable interaction structure is approximately serializable---an assumption that holds for the block and chain-structured diagnostics in our suite but not for the dense rotated control function.

\subsection{Biological Substrate: Chromatin Loop Extrusion}

Rather than loosely associating behavioral rules with a biological metaphor, LEL abstracts a specific structural idea from the biophysics of three-dimensional genome architecture. In eukaryotes, the linear DNA fiber is compacted and partitioned by active \emph{loop extrusion} driven primarily by condensin and cohesin motor proteins (SMC complexes) \cite{Davidson2021, Mirny2021}. These proteins bind to the chromosome and processively extrude the one-dimensional chain into loops, constrained by orientation-specific barriers such as CTCF binding sites \cite{Davidson2023_CTCF, Dequeker2022} and context-dependent chromatin states \cite{Banigan2020, Sept2025, Banigan2023}. Condensin complexes can traverse past one another, producing nested or overlapping loop structures \cite{Kim2020_bypass, Higashi2022, Corsi2023}. The result is a partitioning of the one-dimensional genome into overlapping topological domains governed by local, probabilistic boundary signals.

LEL borrows the abstract idea that an ordered sequence combined with locally responsive boundaries can support overlapping domain discovery. The specific algorithmic choices---the interaction scoring rule, the sigmoid barrier law, the utility-based collision resolution---are engineering decisions made by the authors. They are not direct consequences of the biology, and the biological analogy does not constitute evidence for their optimality. The analogy serves as a source of design intuition, not as a proof of correctness.

\section{Algorithm Design: Loop-Extrusion Linkage}
\label{sec:method}

The Loop-Extrusion Linkage (LEL) operator wraps a standard variation strategy with a structure-learning layer that discovers and exploits variable interactions. It performs four sequential operations: (1) online interaction-graph construction, (2) spectral variable ordering, (3) probabilistic interval generation with adaptive barriers, and (4) blockwise variation on the discovered subsets. The underlying search operator is a conventional DE/rand/1/bin mutation with elite attraction, deliberately chosen to be standard so that any performance differences are attributable to the structural discovery mechanism. We describe each phase below with sufficient detail for independent reimplementation.

\subsection{Phase 1: Interaction Graph Estimation}

LEL maintains a dense interaction score matrix $W \in \mathbb{R}^{d \times d}$ recording pairwise variable relationships inferred from successful optimization steps. We use ``interaction score'' rather than ``covariance'' to emphasize that $W$ is not a statistical covariance matrix---it is an asymmetric, non-negative accumulator that is later symmetrized before spectral analysis.

After each step that yields an improvement $\Delta f < 0$ (minimization), LEL identifies the \emph{active support} $\mathcal{A} = \{i : |\Delta x_i| > \tau\}$ where $\tau = 10^{-8}$. The improvement step is stored in a rolling archive of the most recent $N_{\text{arch}} = 200$ successful steps. From this archive, LEL computes the weighted absolute correlation among active coordinates, where each archive entry is weighted by its improvement magnitude $|\Delta f|$. The interaction score update is an exponential moving average:
\begin{equation}
W_{uv}^{(t)} = (1 - \rho) \, W_{uv}^{(t-1)} + \rho \, \bigl|\operatorname{corr}_w(\Delta x_u, \Delta x_v)\bigr|
\end{equation}
where $\rho = 0.3$ is the decay rate and $\operatorname{corr}_w$ denotes the weighted Pearson correlation computed from the archive entries using improvement-magnitude weights. Self-interactions are set to zero: $W_{ii} = 0$ for all $i$. Variables that are outside $\mathcal{A}$ in recent steps have their scores decayed at rate $0.1\rho$ toward zero, ensuring that stale interaction estimates do not persist indefinitely.

For the spectral analysis that follows, $W$ is sparsified by retaining only the $k = 10$ strongest neighbors per variable, then symmetrized:
\begin{equation}
A = (W_{\text{sparse}} + W_{\text{sparse}}^T) / 2
\end{equation}
This sparsification prevents spurious weak interactions from dominating the spectral ordering.

\textbf{Initialization.} When $W$ is zero or near-zero (fewer than 3 archive entries), no meaningful interaction structure is available. During this warm-up phase, LEL falls back to global DE/rand/1/bin variation, identical in behavior to the base evaluator operating on all $d$ variables simultaneously.

\subsection{Phase 2: Spectral Variable Ordering}

Every $T = 5$ iterations, LEL recomputes a one-dimensional variable ordering via spectral seriation. Given the symmetrized adjacency $A$, the unnormalized graph Laplacian is defined as:
\begin{equation}
L = D - A, \quad D_{ii} = \sum_j A_{ij}
\end{equation}
The Fiedler vector $\mathbf{v}_2$---the eigenvector corresponding to the second-smallest eigenvalue $\lambda_2$ of $L$---is computed using the shift-invert Lanczos method (\texttt{scipy.sparse.linalg.eigsh} with $\sigma = 10^{-10}$, requesting 2 eigenvectors). Variables are then ordered by sorting on $\mathbf{v}_2$:
\begin{equation}
    \pi \gets \operatorname{argsort}(\mathbf{v}_2)
\end{equation}
This places strongly interacting variables in adjacent positions along a linear ordering $\sigma = (\pi(1), \pi(2), \ldots, \pi(d))$. We emphasize that this is a \emph{heuristic} spectral ordering \cite{Atkins1998}, not a provably optimal arrangement for arbitrary interaction graphs. Its quality depends on the degree to which the true interaction structure is approximately serializable---a condition that holds for block and chain structures but not for dense rotated problems.

\textbf{Tie-breaking and degeneracy.} When $\lambda_2$ has multiplicity greater than one (e.g., because $A$ has disconnected components), the Fiedler vector is not uniquely defined. In practice, we use the first eigenvector returned by the iterative solver, which provides a consistent but potentially arbitrary ordering in degenerate cases. Since degenerate Laplacians arise primarily during early iterations (when $W$ is sparse), this ambiguity resolves naturally as interaction estimates accumulate.

For each adjacent pair $(\sigma_r, \sigma_{r+1})$ in the ordering, we compute a \emph{cross-boundary interaction strength} that quantifies how much interaction weight spans boundary $r$:
\begin{equation}
\Gamma_t(r) = \sum_{\substack{u \leq r,\; v > r}} A_{\pi(u),\pi(v)}, \quad r \in \{1,\ldots,d-1\}
\end{equation}
Small values of $\Gamma(r)$ indicate plausible variable-group separators; large values indicate strong coupling that should not be broken by a block boundary.

\subsection{Phase 3: Adaptive Barrier Learning}

LEL maintains a barrier strength vector $\boldsymbol{\beta} \in [0,1]^{d-1}$, initialized at $\beta_r = 0.5$ for all $r$, representing the learned boundary strength at each position in the seriated ordering. Barriers are updated from evidence accumulated in the archive: for each successful step, LEL identifies whether the active coordinates (in ordered position space) span across each boundary $r$, and accumulates weighted evidence counts:
\begin{align}
E_{\text{cross}}(r) &= \sum_{s \in \text{archive}} w_s \, \mathbb{1}[\text{step } s \text{ crosses } r] \\
E_{\text{within}}(r) &= \sum_{s \in \text{archive}} w_s \, \mathbb{1}[\text{step } s \text{ does not cross } r]
\end{align}
where $w_s = |\Delta f_s|$ is the improvement magnitude of archive entry $s$, and a step ``crosses'' boundary $r$ if it has active coordinates on both sides of position $r$ in the seriated ordering.

The barrier update target combines the current barrier value with evidence-driven sigmoid feedback:
\begin{equation}
\beta_r^{(t+1)} = \Pi_{[0,1]}\!\left[(1 - \eta_\beta)\,\beta_r^{(t)} + \eta_\beta \, \sigma\!\left(\kappa \cdot \frac{E_{\text{within}} - E_{\text{cross}}}{E_{\text{within}} + E_{\text{cross}} + \varepsilon}\right)\right]
\end{equation}
where $\eta_\beta = 0.15$ is the barrier learning rate, $\kappa = 3.0$ is the evidence scaling parameter, $\varepsilon = 10^{-10}$ prevents division by zero, $\sigma(\cdot)$ is the logistic sigmoid $\sigma(z) = 1/(1 + e^{-z})$, and $\Pi_{[0,1]}$ denotes projection onto $[0, 1]$. The intuition is straightforward: when historical evidence indicates that successful improvements tend to operate within one side of boundary $r$ rather than across it, the barrier strengthens. When cross-boundary improvements dominate, the barrier weakens.

High $\beta_r$ indicates a strong learned boundary separating positions $r$ and $r{+}1$. The probability of an extrusion loop crossing boundary $r$ during interval generation is then:
\begin{equation}
P(\text{cross}_r) = \sigma\!\left(\alpha\,(\Gamma_t(r) - \beta_r)\right), \quad \alpha = 5.0
\end{equation}
When the cross-boundary interaction $\Gamma(r)$ exceeds the learned barrier $\beta_r$, crossing is likely; when the barrier dominates, crossing is suppressed. This creates a competition between the structural evidence (interaction graph) and the search evidence (barrier learning), allowing domain boundaries to be discovered from optimization dynamics.

\subsection{Phase 4: Stochastic Interval Extrusion}

LEL instantiates $K = \max(4, \lfloor d/8 \rfloor)$ extruders per iteration, spawned at importance-weighted anchor positions along the seriated ordering. The importance weight for position $r$ is proportional to the local interaction density, so anchors are more likely to start in regions of strong variable coupling. Each extruder $k$ starts at a discrete anchor position $v_k \in \{1,\ldots,d\}$ and expands bidirectionally along the linear ordering:

\textbf{Right expansion:} Starting from $r = v_k$, increment $r \gets r+1$ while $r < d$ and $\mathcal{U}(0,1) < P(\text{cross}_r)$ and $r - v_k < L_{\max}$, where $L_{\max} = 24$ is the maximum half-window size.

\textbf{Left expansion:} Starting from $l = v_k$, decrement $l \gets l-1$ while $l \geq 1$ and $\mathcal{U}(0,1) < P(\text{cross}_{l})$ and $v_k - l < L_{\max}$.

The interval boundaries are clamped to $[1, d]$, preventing out-of-range access. The resulting interval $[l_k, r_k]$ maps to the variable subset $B_k = \{\sigma_{l_k}, \ldots, \sigma_{r_k}\}$ in the original coordinate space.

\textbf{Collision resolution.} When intervals from different extruders overlap in the seriated ordering, LEL computes a utility score for each extruder that combines its historical fitness improvement with a size-normalization penalty and boundary cost. Extruders are processed in decreasing utility order: the highest-utility extruder is accepted first, and lower-utility extruders that overlap with already-resolved higher-utility blocks enter a queue. Queued extruders that remain unresolved for $Q_{\max} = 3$ consecutive iterations are shrunk to their non-overlapping portion or reseeded at a new random anchor.

\subsection{Phase 5: Blockwise Variation (Base Evaluator)}

For each resolved block $B_k$, LEL applies variation \emph{only} on the coordinates in $B_k$, holding all other variables fixed at their current values. This blockwise restriction is the mechanism through which the learned structure translates into search behavior: if the ordering and intervals correctly identify groups of interacting variables, then searching within those groups avoids wasting evaluations on irrelevant coordinate combinations.

The variation operator is a weighted combination of three components:
\begin{equation}
y_{B_k} = x_{B_k} + \lambda_1 (\mu_{B_k} - x_{B_k}) + \lambda_2 (x_{p,B_k} - x_{q,B_k}) + \lambda_3 \xi_{B_k}
\end{equation}
where $\mu_{B_k}$ is the mean of the elite fraction ($\tau = 0.3$) of the population projected onto $B_k$, $x_p$ and $x_q$ are two randomly selected population members (providing the differential component), $\xi_{B_k} \sim \mathcal{N}(0, \text{diag}(\sigma^2_{B_k}))$ is Gaussian noise scaled by the coordinate-wise standard deviation of the elite subset, and the weights are $\lambda_1 = 0.4$, $\lambda_2 = 0.5$, $\lambda_3 = 0.1$. The global fallback (used during warm-up and when structure confidence is low) is standard DE/rand/1/bin with adaptive $F \in [0.5, 0.8]$ and $CR = 0.9$.

\subsection{Computational Complexity}

Per iteration, the dominant costs are: (1)~interaction-graph update via weighted correlation over the archive, $O(N_{\text{arch}} \cdot |\mathcal{A}|^2)$ where $|\mathcal{A}|$ is the active support size; (2)~Fiedler vector computation every $T$ iterations via shift-invert Lanczos on the sparse symmetrized adjacency, approximately $O(d \cdot \text{nnz}(L) \cdot n_{\text{iter}})$ where $\text{nnz}$ is the number of nonzero entries; (3)~$K$ extrusion steps, each $O(L_{\max})$; and (4)~$K$ blockwise variation steps, each $O(L_{\max} \cdot N)$ where $N$ is the population size. The interaction matrix $W$ is stored dense at $O(d^2)$; for the present study at $d = 96$, this is approximately 74~KB. For scaling to $d \gg 1000$, sparse-only storage and incremental Laplacian updates would be necessary; we do not claim that the current dense implementation scales to such regimes without modification.

\section{Experimental Evaluation}
\label{sec:evaluation}

This evaluation is designed as a \emph{mechanism study}: we investigate which components of the LEL operator contribute genuine value on synthetic functions with known structural properties. We do \emph{not} claim this constitutes a comprehensive competitive benchmark; rather, it is a targeted diagnostic suite designed to isolate specific structural hypotheses. The functions are chosen to span a range of interaction topologies from fully separable through block-structured to densely coupled, allowing us to map the boundary of the method's useful operating regime.

\subsection{Synthetic Diagnostic Functions}

We evaluate at $d=96$ using six functions, each probing a different structural hypothesis. All functions use bounds $[-5, 5]^d$ and have global optimum $f^* = 0$. The population is initialized uniformly at random within these bounds.

\begin{enumerate}
    \item \textbf{S1: Separable Sphere (Negative control).} No variable interactions; structural operators are unnecessary. This function tests whether LEL's structural machinery introduces harmful overhead when no exploitable structure exists.
    \begin{equation}
        f_1(x) = \sum_{i=1}^{d} x_i^2
    \end{equation}

    \item \textbf{S2: Contiguous-Block Rosenbrock.} Twelve non-overlapping 8-variable Rosenbrock blocks in natural coordinate order:
    \begin{equation}
        f_2(x) = \sum_{k=0}^{11} R\bigl(x_{8k+1}, \ldots, x_{8(k+1)}\bigr)
    \end{equation}
    where $R(z) = \sum_{i=1}^{|z|-1}[100(z_{i+1}-z_i^2)^2 + (1-z_i)^2]$. This tests whether the method can identify and exploit clean block structure when it aligns with coordinate order.

    \item \textbf{S3: Permuted-Block Rosenbrock.} Same block structure as S2, but applied through a fixed random permutation $P$ generated with seed 123: $f_3(x) = f_2(P^{-1}x)$. This is the most informative function for the seriation hypothesis, as it tests whether the learned ordering can recover hidden block structure in the absence of trivial coordinate adjacency.

    \item \textbf{S4: Overlapping-Window Rosenbrock.} Sliding windows of size 8 with stride 4, producing 23 overlapping Rosenbrock blocks:
    \begin{equation}
        f_4(x) = \sum_{j=0}^{22} R\bigl(x_{4j+1}, \ldots, x_{4j+8}\bigr)
    \end{equation}
    This tests whether LEL's overlap-aware interval search and queueing mechanism outperform non-overlapping decomposition methods on problems with shared variables between groups.

    \item \textbf{S5: Banded Quadratic (Chain interaction).} Tridiagonal coupling with logarithmically increasing diagonal weights:
    \begin{equation}
        f_5(x) = \sum_{i=1}^{d} a_i x_i^2 + \lambda \sum_{i=1}^{d-1} (x_i - x_{i+1})^2
    \end{equation}
    with $a_i = 10^{-1 + 2(i-1)/(d-1)}$ (i.e., \texttt{np.logspace(-1,1,d)}) and $\lambda = 10$. The chain-like nearest-neighbor coupling creates a banded interaction structure that is naturally serializable.

    \item \textbf{S6: Dense Rotated Ellipsoid (Negative control).} Full-rank interaction via random orthogonal rotation:
    \begin{equation}
        f_6(x) = x^T (Q^T D Q) x
    \end{equation}
    where $Q$ is a random orthogonal matrix obtained via QR decomposition of a Gaussian random matrix (seed 456) and $D = \operatorname{diag}(10^{0}, 10^{4/(d-1)}, \ldots, 10^4)$, giving condition number $10^4$. The dense, non-sparse interaction graph renders seriation-based decomposition unhelpful, testing the method's failure mode under structural mismatch.
\end{enumerate}

\subsection{Algorithms and Ablations}

Full LEL uses the complete operator described in Section~\ref{sec:method}, with the blockwise variation operator (elite attraction + DE differential + Gaussian noise) as its base evaluator $E$. The population size is $N = 50$ for all configurations.

\textbf{Baselines.}
\begin{itemize}
    \item \textbf{jSO} \cite{Brest2017}: Success-history adaptive DE with linear population size reduction and current-to-$p$best/1 mutation. Represents a strong single-population metaheuristic that adapts control parameters online without structural decomposition. This is a consistently competitive algorithm in CEC benchmark competitions.
    \item \textbf{DG2+CC} \cite{Omidvar2017}: Differential Grouping version 2 for variable decomposition, followed by Cooperative Coevolution using SaNSDE within each identified group. Represents the state of the art in explicit variable grouping for large-scale continuous optimization. DG2 identifies groups via finite-difference interaction detection and is considered the gold standard for decomposition-based approaches.
\end{itemize}

\textbf{Ablations.} Each ablation removes exactly one LEL component to isolate its contribution:
\begin{itemize}
    \item \textbf{A1 (Graph-Only):} Removes seriation and extrusion entirely; uses deterministic connected-component extraction from the learned interaction graph with edge threshold $\tau = 0.1$. Tests the value of the representation layer (ordering + intervals) over raw graph decomposition.
    \item \textbf{A2 (Random-Order):} Replaces spectral seriation with a fixed random ordering (sampled once at initialization from a uniform permutation). Extrusion and barriers operate on this random order. Tests whether a \emph{learned} ordering is better than an arbitrary one.
    \item \textbf{A3 (No-Barriers):} Replaces adaptive barriers with fixed-size non-overlapping windows of size $L_{\max} = 24$ on the seriated ordering. Tests whether the adaptive barrier-learning mechanism adds value over simple uniform windowing.
    \item \textbf{A4 (No-Queueing):} Removes the conflict resolution queue; overlapping extruded blocks are merged by simple set union. Tests whether utility-based priority ordering of overlapping intervals improves performance.
    \item \textbf{A5 (Random-Intervals):} Replaces the learned extrusion process with uniformly random intervals of matched expected length on the seriated ordering. Tests whether the stochastic bidirectional growth mechanism adds value over random interval sampling once a good ordering is available.
\end{itemize}

Not all ablations are tested on all functions. A1 and A2 target the ordering hypothesis (H1), A3 and A5 target the barrier hypothesis (H2), and A4 targets the queueing hypothesis (H3). The negative controls S1 and S6 are tested only against the baselines to characterize the method's behavior under structural mismatch.

\subsection{Experimental Protocol}

Each algorithm--function--budget combination is run with 15 independent seeds (1--15). All algorithms share the same random seed per trial, ensuring that initial populations and any stochastic decisions with shared infrastructure are matched. We report the \textbf{final log-gap}: $\log_{10}(f(x_{\text{best}}) - f^* + \varepsilon)$ with $\varepsilon = 10^{-10}$ (lower is better; negative values indicate solutions within $10^{-10}$ of the optimum).

Statistical significance is assessed via the \textbf{Wilcoxon signed-rank test} \cite{More2009}. This is a paired nonparametric test, appropriate here because runs are \emph{paired by seed}: for each seed $s \in \{1,\ldots,15\}$, the signed difference between the log-gap of algorithm A and Full LEL at seed $s$ forms the paired observation. This pairing controls for seed-dependent variation in initial conditions and random function components (e.g., the permutation in S3, the rotation in S6). Multiple comparisons against Full LEL are controlled by Holm--Bonferroni correction at family-wise $\alpha = 0.05$. We report the \textbf{Vargha--Delaney} $\hat{A}_{12}$ effect size, defined as $\hat{A}_{12} = P(X_{\text{alg}} > X_{\text{LEL}}) + 0.5 \cdot P(X_{\text{alg}} = X_{\text{LEL}})$; values above 0.5 indicate the alternative algorithm has a \emph{larger} (worse) log-gap than Full LEL, while values below 0.5 indicate Full LEL has the worse log-gap.

\textbf{Note on sep-CMA-ES:} A separable CMA-ES baseline was planned but all 180 runs terminated with an implementation error in the population-size interface. These results are excluded rather than reported incompletely. Future work should include this important baseline.

\subsection{Convergence Distributions}

\begin{figure}[htbp]
\centering
\includegraphics[width=0.95\textwidth]{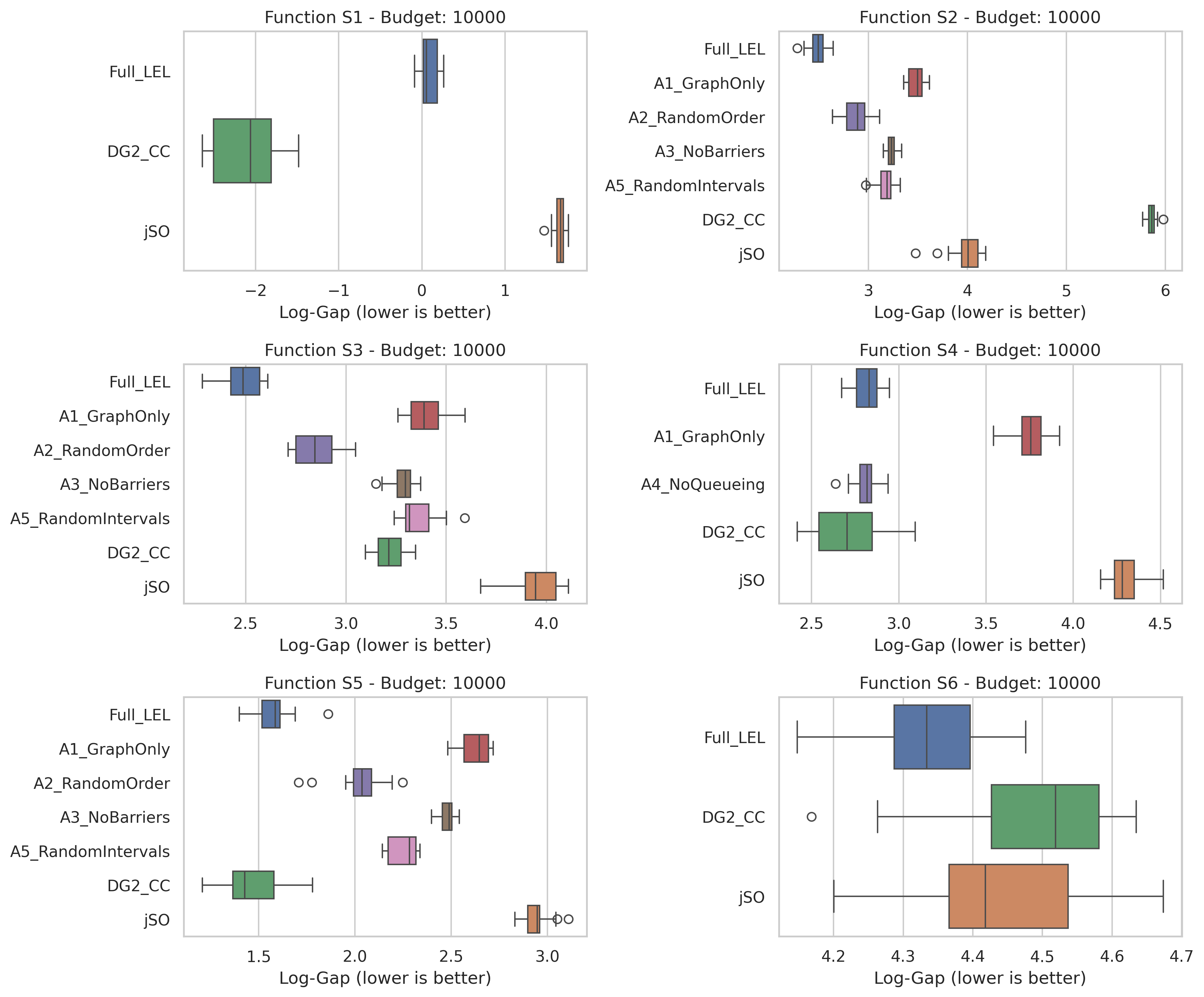}
\caption{Final log-gap distributions at $10^4$ evaluations. Full LEL achieves the best median on S2, S3, and S6, and is competitive on S5. On S1 and S4, DG2+CC achieves the best median.}
\label{fig:boxplots_10k}
\end{figure}

\begin{figure}[htbp]
\centering
\includegraphics[width=0.95\textwidth]{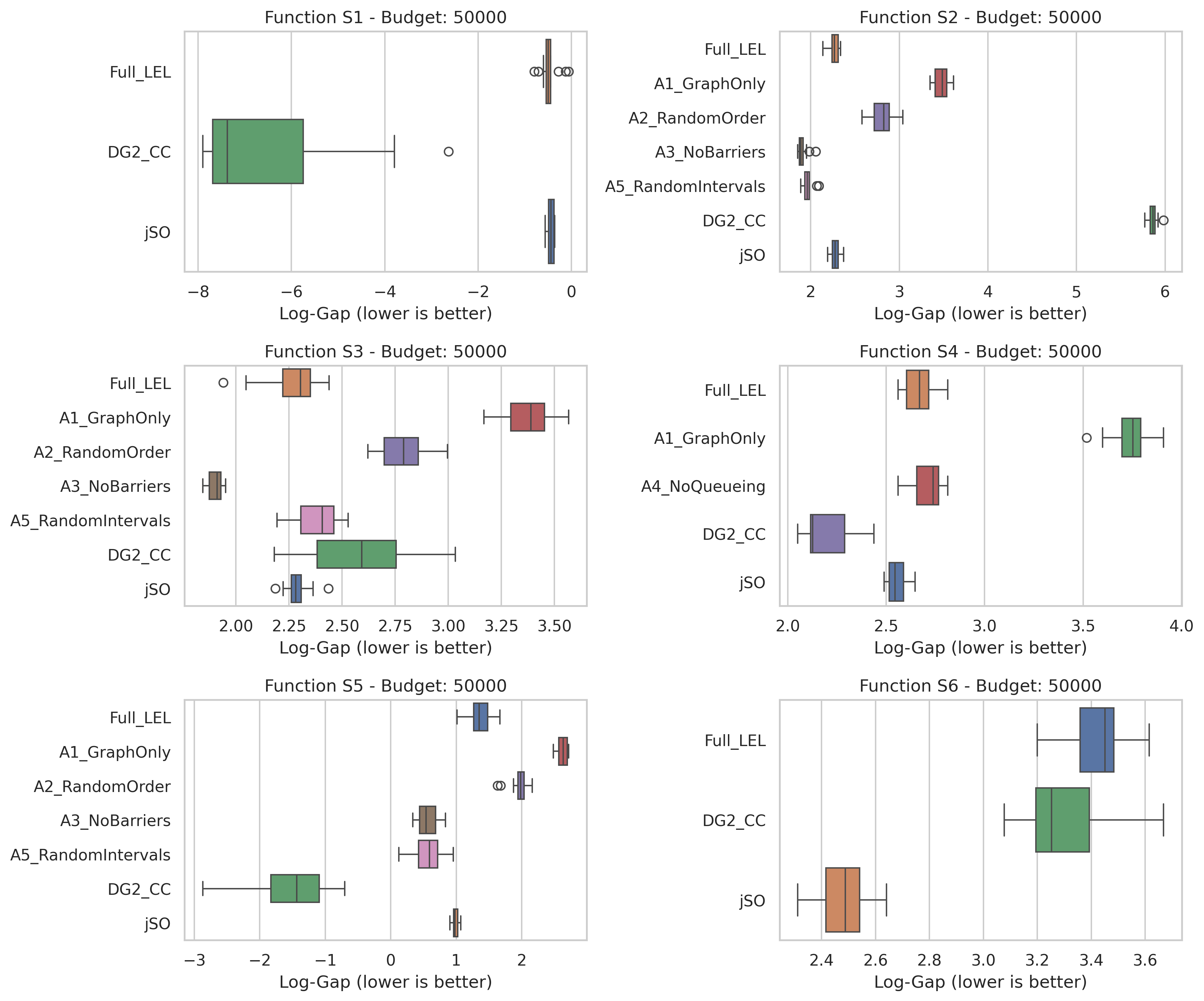}
\caption{Final log-gap distributions at $5 \cdot 10^4$ evaluations. With additional budget, simpler ablations (A3, A5) and baselines (DG2+CC, jSO) often surpass Full LEL, suggesting that the adaptive barrier mechanism may over-constrain late-stage search.}
\label{fig:boxplots_50k}
\end{figure}

Figures~\ref{fig:boxplots_10k} and \ref{fig:boxplots_50k} show the empirical distributions of final log-gap values across the 15 seeds for each algorithm--function combination at both evaluation budgets.

\subsection{Results}

\begin{table}[ht]
\centering
\caption{Synthetic benchmark results ($10^4$ evaluations). Metric: final log-gap (lower is better). Best median per function in \textbf{bold}. Significance stars: {*}{*}{*}\,$p{<}0.001$, {*}{*}\,$p{<}0.01$, {*}\,$p{<}0.05$ (Wilcoxon signed-rank vs.\ Full\_LEL, Holm--Bonferroni corrected). $\hat{A}_{12}$: probability that the row algorithm produces a worse log-gap than Full LEL.}
\label{tab:results_10k}
\resizebox{\textwidth}{!}{%
\begin{tabular}{@{}llrrrr@{}}
\toprule
\textbf{Function} & \textbf{Algorithm} & \textbf{Med LogGap} & \textbf{IQR} & \textbf{$\hat{A}_{12}$ vs LEL} & \textbf{$p$-value} \\
\midrule
\multirow{3}{*}{\textbf{S1: Sep.\ Sphere}}
& \textbf{DG2\_CC} & \textbf{-2.0627} & 0.6948 & 0.00 & ${<}0.001$*** \\
& Full\_LEL & 0.0532 & 0.1642 & - & - \\
& jSO & 1.6662 & 0.0811 & 1.00 & ${<}0.001$*** \\
\midrule
\multirow{7}{*}{\textbf{S2: Contig.\ Rosen.}}
& A1\_GraphOnly & 3.4968 & 0.1330 & 1.00 & ${<}0.001$*** \\
& A2\_RandomOrder & 2.8895 & 0.1817 & 0.99 & ${<}0.001$*** \\
& A3\_NoBarriers & 3.2311 & 0.0581 & 1.00 & ${<}0.001$*** \\
& A5\_RandIntervals & 3.1859 & 0.0997 & 1.00 & ${<}0.001$*** \\
& DG2\_CC & 5.8642 & 0.0534 & 1.00 & ${<}0.001$*** \\
& \textbf{Full\_LEL} & \textbf{2.4920} & 0.1045 & - & - \\
& jSO & 4.0057 & 0.1636 & 1.00 & ${<}0.001$*** \\
\midrule
\multirow{7}{*}{\textbf{S3: Perm.\ Rosen.}}
& A1\_GraphOnly & 3.3890 & 0.1354 & 1.00 & ${<}0.001$*** \\
& A2\_RandomOrder & 2.8452 & 0.1789 & 1.00 & ${<}0.001$*** \\
& A3\_NoBarriers & 3.2962 & 0.0659 & 1.00 & ${<}0.001$*** \\
& A5\_RandIntervals & 3.3172 & 0.1161 & 1.00 & ${<}0.001$*** \\
& DG2\_CC & 3.2135 & 0.1122 & 1.00 & ${<}0.001$*** \\
& \textbf{Full\_LEL} & \textbf{2.4875} & 0.1438 & - & - \\
& jSO & 3.9456 & 0.1513 & 1.00 & ${<}0.001$*** \\
\midrule
\multirow{5}{*}{\textbf{S4: Overlap Win.}}
& A1\_GraphOnly & 3.7570 & 0.1092 & 1.00 & ${<}0.001$*** \\
& A4\_NoQueueing & 2.8188 & 0.0668 & 0.46 & 0.804 \\
& \textbf{DG2\_CC} & \textbf{2.7029} & 0.3051 & 0.25 & 0.035* \\
& Full\_LEL & 2.8305 & 0.1166 & - & - \\
& jSO & 4.2817 & 0.1133 & 1.00 & ${<}0.001$*** \\
\midrule
\multirow{7}{*}{\textbf{S5: Banded Quad.}}
& A1\_GraphOnly & 2.6460 & 0.1275 & 1.00 & ${<}0.001$*** \\
& A2\_RandomOrder & 2.0368 & 0.0922 & 0.99 & ${<}0.001$*** \\
& A3\_NoBarriers & 2.4902 & 0.0506 & 1.00 & ${<}0.001$*** \\
& A5\_RandIntervals & 2.2843 & 0.1444 & 1.00 & ${<}0.001$*** \\
& \textbf{DG2\_CC} & \textbf{1.4282} & 0.2136 & 0.25 & 0.055 \\
& Full\_LEL & 1.5867 & 0.0936 & - & - \\
& jSO & 2.9482 & 0.0611 & 1.00 & ${<}0.001$*** \\
\midrule
\multirow{3}{*}{\textbf{S6: Dense Ellip.}}
& DG2\_CC & 4.5185 & 0.1547 & 0.82 & 0.022* \\
& \textbf{Full\_LEL} & \textbf{4.3336} & 0.1092 & - & - \\
& jSO & 4.4180 & 0.1711 & 0.78 & 0.022* \\
\bottomrule
\end{tabular}%
}
\end{table}

\begin{table}[ht]
\centering
\caption{Synthetic benchmark results ($5 \cdot 10^4$ evaluations). Metric: final log-gap (lower is better). Best median per function in \textbf{bold}. Significance stars: {*}{*}{*}\,$p{<}0.001$, {*}{*}\,$p{<}0.01$, {*}\,$p{<}0.05$ (Wilcoxon signed-rank vs.\ Full\_LEL, Holm--Bonferroni corrected). $\hat{A}_{12}$: probability that the row algorithm produces a worse log-gap than Full LEL.}
\label{tab:results_50k}
\resizebox{\textwidth}{!}{%
\begin{tabular}{@{}llrrrr@{}}
\toprule
\textbf{Function} & \textbf{Algorithm} & \textbf{Med LogGap} & \textbf{IQR} & \textbf{$\hat{A}_{12}$ vs LEL} & \textbf{$p$-value} \\
\midrule
\multirow{3}{*}{\textbf{S1: Sep.\ Sphere}}
& \textbf{DG2\_CC} & \textbf{-7.3724} & 1.9388 & 0.00 & ${<}0.001$*** \\
& Full\_LEL & -0.4979 & 0.0959 & - & - \\
& jSO & -0.4425 & 0.1165 & 0.62 & 0.679 \\
\midrule
\multirow{7}{*}{\textbf{S2: Contig.\ Rosen.}}
& A1\_GraphOnly & 3.4868 & 0.1309 & 1.00 & ${<}0.001$*** \\
& A2\_RandomOrder & 2.8240 & 0.1709 & 1.00 & ${<}0.001$*** \\
& \textbf{A3\_NoBarriers} & \textbf{1.8876} & 0.0442 & 0.00 & ${<}0.001$*** \\
& A5\_RandIntervals & 1.9593 & 0.0466 & 0.00 & ${<}0.001$*** \\
& DG2\_CC & 5.8641 & 0.0534 & 1.00 & ${<}0.001$*** \\
& Full\_LEL & 2.2668 & 0.0703 & - & - \\
& jSO & 2.2752 & 0.0676 & 0.53 & 0.934 \\
\midrule
\multirow{7}{*}{\textbf{S3: Perm.\ Rosen.}}
& A1\_GraphOnly & 3.3883 & 0.1579 & 1.00 & ${<}0.001$*** \\
& A2\_RandomOrder & 2.7898 & 0.1602 & 1.00 & ${<}0.001$*** \\
& \textbf{A3\_NoBarriers} & \textbf{1.9114} & 0.0555 & 0.01 & ${<}0.001$*** \\
& A5\_RandIntervals & 2.4065 & 0.1554 & 0.76 & 0.018* \\
& DG2\_CC & 2.5928 & 0.3715 & 0.86 & ${<}0.001$*** \\
& Full\_LEL & 2.3032 & 0.1313 & - & - \\
& jSO & 2.2808 & 0.0464 & 0.48 & 0.978 \\
\midrule
\multirow{5}{*}{\textbf{S4: Overlap Win.}}
& A1\_GraphOnly & 3.7530 & 0.0959 & 1.00 & ${<}0.001$*** \\
& A4\_NoQueueing & 2.7384 & 0.1118 & 0.67 & 0.208 \\
& \textbf{DG2\_CC} & \textbf{2.1266} & 0.1738 & 0.00 & ${<}0.001$*** \\
& Full\_LEL & 2.6695 & 0.1123 & - & - \\
& jSO & 2.5455 & 0.0715 & 0.12 & ${<}0.001$*** \\
\midrule
\multirow{7}{*}{\textbf{S5: Banded Quad.}}
& A1\_GraphOnly & 2.6340 & 0.1278 & 1.00 & ${<}0.001$*** \\
& A2\_RandomOrder & 1.9840 & 0.0972 & 1.00 & ${<}0.001$*** \\
& A3\_NoBarriers & 0.5402 & 0.2430 & 0.00 & ${<}0.001$*** \\
& A5\_RandIntervals & 0.5933 & 0.2914 & 0.00 & ${<}0.001$*** \\
& \textbf{DG2\_CC} & \textbf{-1.4345} & 0.7410 & 0.00 & ${<}0.001$*** \\
& Full\_LEL & 1.3472 & 0.2133 & - & - \\
& jSO & 0.9800 & 0.0607 & 0.02 & ${<}0.001$*** \\
\midrule
\multirow{3}{*}{\textbf{S6: Dense Ellip.}}
& DG2\_CC & 3.2532 & 0.1978 & 0.28 & 0.083 \\
& Full\_LEL & 3.4509 & 0.1246 & - & - \\
& \textbf{jSO} & \textbf{2.4879} & 0.1253 & 0.00 & ${<}0.001$*** \\
\bottomrule
\end{tabular}%
}
\end{table}

\subsection{Discussion}

We organize the discussion around four structural hypotheses derived from the method's design.

\subsubsection{H1: Does learned spectral ordering improve over graph-only and random ordering?}

\textbf{Supported.} On S2 and S3 at $10^4$ evaluations, Full LEL achieves the best median log-gap (2.49 and 2.49, respectively), significantly outperforming all ablations and both baselines ($p < 0.001$, $\hat{A}_{12} \geq 0.99$ for all pairwise comparisons). The margin is substantial. A1 (Graph-Only) stagnates near 3.5 on both functions and shows negligible improvement from $10^4$ to $5 \cdot 10^4$ (e.g., 3.50 $\to$ 3.49 on S2), demonstrating that an interaction graph alone, even when learned online, is insufficient for effective optimization---it must be converted into an actionable variable arrangement that the search operator can exploit. A2 (Random-Order) achieves 2.89 and 2.85 at $10^4$, consistently worse than the learned ordering by 0.36--0.40 log units. On a logarithmic scale, this corresponds to roughly $2$--$2.5\times$ larger optimality gaps, indicating that the quality of the ordering matters, not merely the fact that interval-based search is used.

S3 is particularly informative because the natural coordinate order has been destroyed by a random permutation: the 12 Rosenbrock blocks are scattered across the 96-dimensional coordinate space. The fact that Full LEL recovers useful structure from estimated interactions alone---achieving a median log-gap comparable to S2 despite the permutation---provides the strongest evidence that the spectral seriation step performs meaningful order recovery rather than simply benefiting from coincidental coordinate adjacency.

On S5 (banded quadratic), Full LEL also substantially outperforms A1 and A2 at both budgets (by 1.06 and 0.45 log units at $10^4$, respectively), confirming that the learned ordering captures chain-like nearest-neighbor coupling structure.

\subsubsection{H2: Do adaptive barriers improve performance?}

\textbf{Mixed evidence.} At $10^4$ evaluations, Full LEL outperforms A3 (No-Barriers, which uses fixed-size windows of $L_{\max} = 24$) on all structured functions, with the advantage being strongest on S2 (2.49 vs.\ 3.23, a 0.74 log-unit gap) and S3 (2.49 vs.\ 3.30, a 0.81 log-unit gap). This is consistent with the hypothesis that at low budgets, adaptive barriers help discover block boundaries more efficiently than blind fixed-size windowing. When the optimization budget is scarce, the barrier mechanism appears to focus evaluations on correctly sized subsets sooner.

However, at $5 \cdot 10^4$ evaluations, the picture reverses substantially. A3 achieves the best median on S2 ($1.89$ vs.\ $2.27$ for Full LEL) and S3 ($1.91$ vs.\ $2.30$), and dramatically outperforms Full LEL on S5 ($0.54$ vs.\ $1.35$, a 0.81 log-unit advantage favoring the no-barrier ablation). A5 (Random-Intervals) shows a similar pattern, reaching $1.96$ on S2 and $0.59$ on S5 at $5 \cdot 10^4$. These results suggest that when sufficient budget is available for the fixed windows to explore thoroughly, the adaptive barrier mechanism may over-constrain late-stage search. One possible explanation is that barriers calibrated during early exploration---when the interaction graph is noisy and incomplete---persist at strengths that are no longer appropriate once the function landscape has been more fully sampled, effectively locking the method into suboptimal domain boundaries.

\subsubsection{H3: Does queue-based conflict resolution improve on overlapping structure?}

\textbf{Weak evidence.} On S4 (overlapping windows), Full LEL and A4 (No-Queueing) produce statistically indistinguishable results at $10^4$ ($p = 0.804$, $\hat{A}_{12} = 0.46$) and at $5 \cdot 10^4$ ($p = 0.208$, $\hat{A}_{12} = 0.67$). Both are substantially better than A1 (Graph-Only), which stagnates near 3.75 at both budgets, confirming that interval-based search helps on overlapping structure. However, the queueing mechanism itself does not produce a statistically significant benefit at either budget under the current parameter settings.

It is worth noting that DG2+CC achieves the best median on S4 at both budgets ($2.70$ at $10^4$ and $2.13$ at $5 \cdot 10^4$), suggesting that DG2's finite-difference grouping, despite producing non-overlapping groups, provides a more effective decomposition than LEL's overlap-aware but less precise interval-based approach on this particular function.

\subsubsection{H4: How does LEL behave when its inductive bias is mismatched?}

\textbf{Graceful degradation.} On S1 (separable sphere), Full LEL achieves a median log-gap of $-0.50$ at $5 \cdot 10^4$, which is comparable to jSO ($-0.44$, $p = 0.679$) but far behind DG2+CC ($-7.37$, $p < 0.001$). The structural machinery adds overhead---evaluations are spent learning an interaction graph that contains no meaningful signal---but it does not catastrophically harm performance on a fully separable landscape. DG2+CC excels here because its finite-difference grouping correctly identifies all 96 variables as independently separable and can optimize each one individually.

On S6 (dense rotated ellipsoid), Full LEL achieves the best median at $10^4$ ($4.33$ vs.\ jSO $4.42$, $p = 0.022$), suggesting that even on densely coupled problems, the structured local search provides some early-stage benefit. However, at $5 \cdot 10^4$, jSO wins decisively ($2.49$ vs.\ $3.45$, $p < 0.001$), and DG2+CC also surpasses Full LEL ($3.25$ vs.\ $3.45$, $p = 0.083$). This indicates that the one-dimensional serial-order inductive bias becomes limiting when the true interaction structure is dense and non-serializable: jSO's adaptive global search, unconstrained by structural assumptions, refines more effectively at larger budgets.

\subsubsection{Budget sensitivity}

A notable cross-cutting finding is that Full LEL's relative standing is \emph{budget-dependent}: it achieves the best median on 3 of 6 functions at $10^4$ evaluations but on 0 of 6 at $5 \cdot 10^4$. This budget sensitivity is not unique to LEL; the benchmarking literature has documented that algorithm rankings frequently change across evaluation horizons, and that conclusions drawn at a single budget can be misleading \cite{Piotrowski2025, Vermetten2024}. For LEL specifically, the pattern suggests that the method may be most valuable as an \emph{early-budget structural operator}---it appears to discover exploitable variable groupings rapidly, providing a substantial initial advantage, but its adaptive control mechanisms (barriers and queueing) may plateau or interfere with refinement at larger budgets where simpler mechanisms have enough evaluations to converge through exhaustive local search.

This interpretation is supported by the observation that LEL uniformly outperforms jSO at $10^4$ evaluations across all six functions---including the negative controls---but this advantage erodes or reverses on 4 of 6 functions by $5 \cdot 10^4$. The most promising practical implication is that LEL may function well as a warm-start or initial-phase operator within a hybrid optimization pipeline that switches to less constrained search at later stages.

\subsection{Limitations}

\begin{itemize}
    \item \textbf{Narrow benchmark scope.} Six synthetic functions at one dimensionality ($d = 96$) do not substitute for broad-suite evaluation on established platforms such as COCO/BBOB \cite{Hansen2021}. The results support focused mechanism hypotheses but not general competitiveness claims.
    \item \textbf{No runtime analysis.} We report only function evaluations, not wall-clock time. The overhead of interaction-graph maintenance, spectral ordering, and extrusion logic is not measured; practical competitiveness depends on this overhead being small relative to the cost of objective-function evaluations.
    \item \textbf{Missing baseline.} The planned sep-CMA-ES comparison failed due to an implementation error and is not included. This baseline would be particularly informative on S1 and S6.
    \item \textbf{Single dimensionality.} Scaling behavior to $d \gg 96$ is not evaluated. The dense interaction matrix $W$ and the full Laplacian eigenvector computation may become bottlenecks at higher dimensions.
    \item \textbf{Barrier mechanism not validated.} Current results do not establish that adaptive barriers are necessary; simpler fixed-size windows often perform comparably or better at larger budgets, and the barrier-learning dynamics may benefit from decay or phase-switching mechanisms not explored here.
    \item \textbf{No adversarial graph structures.} The diagnostic suite does not include functions whose interaction topology is fundamentally non-serializable yet sparse (e.g., tree, grid, or hub-and-spoke interaction graphs). Testing against such structures would better delineate the boundary of the seriation-based approach.
\end{itemize}

\section{Conclusion}
\label{sec:conclusion}

We introduced the Loop-Extrusion Linkage (LEL) operator, a structure-learning wrapper for continuous optimization that combines online interaction-graph estimation, spectral variable ordering via the Fiedler vector, adaptive boundary learning, and stochastic interval-based subspace search. Through a targeted synthetic diagnostic study at $d = 96$, we evaluated four structural hypotheses using systematic ablation and comparison against jSO and DG2+CC across two evaluation budgets.

\textbf{Main findings.} The strongest supported result is that learned spectral ordering consistently and substantially improves over graph-only grouping and random variable ordering on problems with block or chain-like interaction structure. On the permuted-block Rosenbrock function (S3)---where the natural coordinate order is destroyed by a random permutation---Full LEL significantly outperforms all ablations and both baselines at $10^4$ evaluations ($p < 0.001$ for all comparisons), demonstrating that the interaction-graph seriation mechanism recovers a useful variable ordering from search history alone. The graph-only ablation (A1) stagnates regardless of budget, confirming that an interaction graph must be converted into an actionable representation before search can exploit it.

However, the adaptive barrier mechanism shows mixed evidence: it helps at low budgets but is often surpassed at higher budgets by simpler fixed-size windowing. At $5 \cdot 10^4$ evaluations, the no-barrier ablation achieves the best median on both block-structured Rosenbrock functions and substantially outperforms Full LEL on the banded quadratic function. This suggests that barrier strengths calibrated during early noisy exploration may persist at suboptimal levels during later refinement. The queueing mechanism does not produce statistically significant benefits on the overlapping-window diagnostic with current parameter settings.

\textbf{Budget sensitivity.} A cross-cutting observation is that Full LEL's competitive standing is strongly budget-dependent, winning 3 of 6 functions at $10^4$ evaluations but none at $5 \cdot 10^4$. This pattern, consistent with the broader benchmarking literature's finding that algorithm rankings are evaluation-horizon dependent \cite{Piotrowski2025, Vermetten2024}, suggests the method may be most valuable as an early-stage structural operator---rapidly discovering exploitable variable groupings---rather than as a long-run whole-optimization policy. A practical direction would be to deploy LEL as a warm-start module within a hybrid pipeline that transitions to unconstrained adaptive search once structural discovery plateaus.

\textbf{Scientific positioning.} Consistent with the growing critique that metaheuristic novelty should be defined by algorithmic mechanism rather than metaphorical inspiration \cite{Sorensen2015, CamachoVillalon2026, Molina2025}, we have attempted to evaluate LEL's components transparently rather than presenting the full method as a validated package. The biology of loop extrusion provided useful design intuition---the idea that an ordered substrate with locally responsive, partially permeable boundaries can support overlapping domain discovery---but the algorithmic evidence must stand on its own terms. The current results support the ordering idea and identify design weaknesses in the control layer.

\textbf{Future work.} Several directions are warranted: (1)~evaluating on standard benchmark suites (COCO/BBOB, CEC large-scale) to establish broader applicability beyond custom diagnostics; (2)~diagnosing and resolving the late-budget erosion of barrier effectiveness, potentially through adaptive barrier decay, annealing schedules, or phase-switching control policies that reduce barrier influence as the evaluation budget grows; (3)~extending to higher dimensionalities ($d \geq 500$) with sparse interaction graph representations and incremental Laplacian updates; (4)~investigating alternative ordering heuristics beyond Fiedler seriation, such as Cuthill--McKee reordering or minimum linear arrangement methods, to determine whether the spectral approach is uniquely effective or whether any reasonable order-recovery procedure suffices; (5)~testing on functions with non-serializable sparse interaction topologies (trees, grids, hub-and-spoke graphs) to better delineate the method's failure boundary; and (6)~incorporating sep-CMA-ES and additional modern baselines in future experimental campaigns.

\bibliographystyle{plain}
\bibliography{references}

\end{document}